# Assessing Classical Machine Learning and Transformer-based Approaches for Detecting AI-Generated Research Text

**Sharanya Parimanoharan**

(University of Peradeniya, Peradeniya, Srilanka

https://orcid.org/0000-0000-0000-0000, sharamano10@gmail.com)

**Ruwan D. Nawarathna**

(University of Peradeniya, Peradeniya, Srilanka

https://orcid.org/0000-0000-0000-0000, ruwand@sci.pdn.ac.lk)

**Abstract:** The rapid adoption of large language models (LLMs) such as ChatGPT has blurred the line between human and AI-generated texts, raising urgent questions about academic integrity, intellectual property, and the spread of misinformation. Thus, reliable AI-text detection is needed for fair assessment to safeguard human authenticity and cultivate trust in digital communication. In this study, we investigate how well current machine learning (ML) approaches can distinguish ChatGPT-3.5-generated texts from human-written texts employing a labeled data set of 250 pairs of abstracts from a wide range of research topics. We test and compare both classical (Logistic Regression armed with classical Bag-of-Words / POS / TF-IDF features) and transformer-based (BERT augmented with N-grams, DistilBERT, BERT with a lightweight custom classifier, and LSTM-based N-gram models) ML detection techniques. As we aim to assess each model's performance in detecting AI-generated research texts, we also aim to test whether an ensemble of these models can outperform any single detector. Results show DistilBERT achieves the overall best performance, while Logistic Regression and BERT-Custom offer solid, balanced alternatives; LSTM- and BERT-N-gram approaches lag. The max voting ensemble of the three best models fails to surpass DistilBERT itself, highlighting the primacy of a single transformer-based representation over mere model diversity. By comprehensively assessing the strengths and weaknesses of these AI-text detection approaches, this work lays a foundation for more robust, transformer frameworks with larger, richer datasets to keep pace with ever-improving generative AI models.

## 1 Introduction

In recent years, the rapid advancements in generative Artificial Intelligence (gen-AI) and Natural Language Processing (NLP) have led to the widespread emergence of AI-generated text across various forms of media and literature [Kreps et al. 2022, Woo et al. 2023, Partadiredja et al. 2020, Amirjalili et al. 2024]. The significance of this evolution has been remarked by the dominance of models like the GPT series that can accurately mimic human languages. Such models generate prompt text responses that are not only contextually appropriate but also highly coherent, making them indistinguishable from original human-generated texts. The ability of such models to generate text that resembles human language has a profound impact on human creativity and learning in various avenues [Lo et al. 2023, Wei et al. 2024, Wang et al. 2023]. While this evolution introduces exciting possibilities to advance digital communication and AI-driven technologies, it also prompts a critical question: How can we eventually differentiate between text crafted by AI and that written by humans?. To address this question, one needs to understand the steps and processes involved during AI-text generation.

AI-text generation essentially means an AI system producing written content, imitating human language patterns and styles. This process relies on algorithms and language models to analyze the input data to generate corresponding output text. As illustrated in Figure 1, a model like GPT-3.5 learns fundamentals of language by analyzing extensive text data from the internet, aiming to understand how words and phrases are used [Ouyang et al. 2022]. Additionally, it also receives feedback from human reviewers, who evaluate and rate the quality of the responses it generates. Then, based on that feedback, the model fine-tunes its parameters to further improve the quality of its output, consequently strengthening the alignment between humans' input text and AI's output text. This multi-level iterative training process not only obscures the difference between the AI-generated and human-written texts but also increases the complexity of the detection process. For instance, models such as GPT-3.5 can even adapt to different writing styles making it challenging to develop fixed rules for detection. They continuously learn from new data, which continuously enhances their versatility and adaptability. In summary, as AI models evolve rapidly, they often surpass the capabilities of current detection methods easily making once-effective approaches outdated.

To effectively detect AI-generated text, we need ML models that capture both surface-level stylistic cues and deep semantic patterns. These ML models can be of two types: first, a classical classifier [Faouzi et al. 2023, Nusinovici et al. 2020] that ingests engineered features like TF-IDF, POS counts, and readability scores, and second, fine-tuned transformer models [Amatriain et al. 2023] whose embeddings capture deeper syntactic and discourse cues in the text. Hybrid or ensemble schemes can also be used to merge these approaches to boost robustness. The majority of the previous studies on AI-generated text detection tend to take distinct approaches: some rely on classical statistical [Gehrmann et al. 2019, Wu et al. 2023] or ML techniques [Alamleh et al. 2023], while others employ transformer-based DL, such as fine-tuned transformer models [Lai et al. 2024, Antoun et al. 2023]. However, only a limited studies compare classical models versus modern transformers under identical conditions [Mikros et al. 2023, Hazim et al. 2024].

In this context, this work unfolds a systematic investigation by attentively testing a range of approaches, including both classical ML models such as Logistic Regression, LSTM neural networks as well as modern transformer models such as BERT and the pretrained DistilBert to detect the AI-generated research text from human-generated versions. These models were selected for their individual strengths and compatibility with data available as well as computational resources. DistilBERT and RoBERTa both originate from the encoder-only BERT framework, yet RoBERTa typically demands a far larger pre-training corpus and greater computational resources. To keep the study lean while still assessing a BERT-style model, we selected DistilBERT which is the compact, compressed sibling of BERT.

Our work is one of the few to combine traditional ML models with modern transformer-based models in a unified framework for AI-generated text detection. Another key feature in our approach is the use of an ensemble max-voting algorithm combining the three best-performing models. Ensemble methods have been explored only sparingly in AI-generated text detection research [Mikros et al. 2023, Alhayan et al. 2024, Abburi et al. 2025]. We used a simple yet effective max voting ensemble: each of the three best models votes on whether a given abstract is AI- or human-written, and the majority decision is taken as the output. Furthermore, we use a unique dataset of 250 research paper abstracts spanning multiple disciplines, contrasting with the majority of prior works that often concentrate on specific domains or languages. Because a detection pipeline requires a balanced dataset of human and AI-authored texts, diversity in topic, length, and generator versions is essential to avoid dataset bias.

Our study finally plays out a comparative analysis of these different ML methods for AI-generated text detection. Through systematic testing and experimentation, we aim to identify the most robust approach to differentiate between AI-generated and human-written text, offering subtle insights into tools, techniques, and implications surrounding AI-generated text detection. Our study will pave the way to uncover the existing complexities, explore strategies, and contribute to the development of effective AI-content detection tools.

The paper is structured in the following order: Firstly, we introduce the problem of AI-generated text detection highlighting its importance and challenges. Secondly, we present a comprehensive literature review discussing various approaches used for AI-generated text detection while highlighting their inherent challenges. Thirdly, we present the methodology we have adopted to carry out our study on comparing various detection approaches. Then we present the results of the study, comparing the effectiveness of the models we used for detecting the AI-generated text and finally we conclude by providing insights into future work that could potentially improve our efforts to detect AI-generated text.

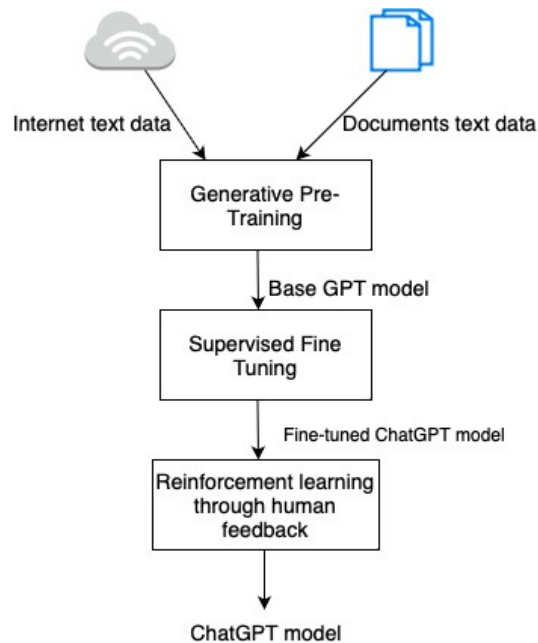

*Figure 1: Process for GPT 3.5 Model Training*

## 2 Literature Review on AI-Generated Text Detection Approaches

AI text detection using ML approaches ranges from classical ML models to sophisticated deep learning (DL) models and transformer-based models. In this section, we aim to provide a comprehensive understanding of the advancements, challenges, and emerging trends with regard to these approaches.

### 2.1 Classical ML for Detection

N-gram language models remain relevant in many NLP tasks despite the emergence of more advanced architectures such as transformers. Meanwhile, researchers also continue to explore various strategies to enhance N-gram models' capabilities [García et al. 2021] through incorporating sub-words for improved word handling as well as using recurrent neural network language modeling for template generation and bidirectional context [Suzuki et al. 2019]. Specifically, these advancements aim to address the limitations of traditional N-gram models and improve their performance in practical applications. Furthermore, these models are based on the principle of predicting the likelihood of a word or token in a sequence, considering the context of the preceding N-1 words. However, they do have limitations, particularly in capturing long-range dependencies and understanding intricate language structures.

In a study [Sarzaeim et al. 2023], a cloud-based prototype is introduced for detecting AI-generated text in scientific writings. The model uses a Multilayer perceptron (MLP) neural network and employs a count vectorizer for text vectorization, distinguishing patterns between human-generated and AI-generated texts. Despite having a smaller training dataset and simplified text pre-processing, the study reported that the proposed model outperformed existing tools like the OpenAI classifier, achieving an accuracy of 89.09%.

### 2.2 DL Models for Detection

In recent works, researchers have integrated the Tunicate Swam Algorithm (TSA) with the Long Short Term Memory Recurrent Neural Network (LSTMRNN) to distinguish both ChatGPT-generated text and human-generated text [Katib et al. 2023]. This method uses techniques such as TF-IDF, word embedding, and count vectorizers to extract important features from text. While the LSTMRNN handles the detection and classification task, the TSA manages parameter selection. This combination is believed to improve the model's ability to detect and understand patterns. However, it also poses some potential challenges. These include concerns about the computing power required, the complexity of explaining decision-making processes, limitations in generalizing across diverse datasets, and the heavy reliance on ensemble models.

Similarly, another study proposes a robust method, SeqXGPT [Wang et al. 2023], which utilizes a sequence labeling approach for detecting AI-generated text at the sentence level. According to the suggested approach, token-wise log probability lists are first extracted from various language models to establish a foundational dataset. Once these foundations are built, SeqXGPT employs convolutional neural networks to extract critical local features, treating word-level log probabilities as indicators of language patterns. Finally, a linear classification layer is trained to analyze contextual features and identify the most common label for each sentence. In a detailed analysis conducted on a custom-built dataset, this model showcased remarkable results. SeqXGPT achieved an impressive F1 score of 98.78%. However, a notable limitation of this approach is the absence of semantic features in the model, hindering its performance. Additionally, further exploration is needed to understand the model's performance on a more diverse dataset.

## 2.3  Transformer-Based Models for Detection

Researchers have also explored unsupervised methods, such as the "DetectGPT" algorithm [Mitchell et al. 2023], which aims to detect machine-generated text without explicit training on labelled data. These methods leverage the inherent differences between human-written and machine-generated text, such as statistical patterns, fluency, and context. The algorithm focuses on analyzing the log-probability functions or curvatures of text passages to differentiate between the two categories. These approaches have the advantage of not needing extensive training data, but they might still require access to model probabilities or other related information. This algorithm relies on the insight that machine-generated text tends to inhabit regions of negative curvature in the log-probability function. By introducing the concept of perturbation discrepancy, the algorithm effectively compares log-probabilities between the original and perturbed versions. In particular, this algorithm has shown strong performance in detecting machine-generated fake news articles.

In a similar study, researchers investigated the characteristics of AI-generated essays compared to human-written essays [Yan et al. 2023]. Researchers generated multiple versions of essays using OpenAI's GPT-3. ETS's e-rater scoring engine was then utilized for feature extraction and scoring, providing valuable insights into the characteristics of both human and AI-generated essays. In this study, two detectors are developed to distinguish between essays. The first detector relied on a fine-tuned Roberta transformer model, known for its effectiveness in text generation and classification. The second detector was built based on e-rater features and an SVM model, incorporating nearly 200 features extracted during e-rater scoring. The performance of both detectors is evaluated using five-fold cross-validation, a robust technique that involves dividing the dataset into five subsets for iterative training and testing. Notably, the results of this research revealed that GPT-3 generated essays closely resemble human-written essays in many aspects, making it challenging for automated scoring systems like e-rater to distinguish. The two detectors mentioned in this study demonstrated high accuracy in detection, exceeding 90%. However, these studies recognize some limitations in their approach: They did not fully explore how people might make changes to AI-generated essays in real life. Additionally, the researchers did not compare the essays generated by a wide range of generative LLMs.

Another study by [Wu and Segura-Bedmar 2025] developed the Giant Language Model Test Room (GLMTR), a visual tool designed to help detect AI-generated text by highlighting words based on how likely they are to be machine-generated using GPT-2 models. Their study evaluated various GPT-2 variants on English and Spanish datasets in the IberLef-AuTexTification 2023 shared task. This study found that the smaller GPT-2 model (gpt2-small) worked best for English texts, scoring 80.19% in accuracy. For Spanish texts, the bigger GPT-2 model (gpt2-xl), even though it wasn't trained on Spanish, did better than the smaller ones with a score of 66.20%. However, GLTR has its own disadvantages: It can sometimes give unclear results and has difficulty detecting more advanced AI-generated text. This shows the challenges in reliably identifying machine-written content with current tools.

## 2.4   AI-Content Detection Online Tools

In a previous study, researchers evaluated the effectiveness of online AI text content detectors such as OpenAI, CrossPlag, Writer, Copyleaks, and GPTZero in differentiating between AI-generated and human-written text [Elkhatat et al. 2023]. OpenAI displayed high sensitivity but low specificity, indicating its capability to identify AI-generated text but facing challenges in recognizing human-generated text. On the other hand, CrossPlag showcased high specificity with human-written text but encountered difficulties with AI-generated text, particularly when detecting text generated by GPT-4. This study highlights significant variability among these detectors, consistently showing high performance in identifying GPT-3.5 generated content compared to GPT-4 generated text and human-written texts. This study concluded that AI detection tools exhibit inconsistency in performance recommending combining AI tools with manual review to ensure a fair evaluation process. Recently, the Ghostbuster platform was proposed using a series of smaller language models for finding patterns in the features and a logistic regression-based classifier for training [Verma et al. 2023]. This method's performance decayed with token counts in the training document data despite showing high generalizability.

The findings, in overall, underscore the dynamic nature of AI evolution, indicating the ongoing challenges with the state-of-the-art online detection tools in adapting to ever-evolving generative AI models.

## 2.5   Challenges in AI-Generated Text Detection

A survey reviewed key methods for detecting AI-generated text, such as watermarking, statistical and stylistic analysis, and ML classifiers [Fraser et al. 2025]. The study emphasized the growing difficulty in distinguishing AI-generated content from human-written text due to rapid advancements in large language models (LLMs). It also identified key challenges, including inconsistent accuracy, dataset limitations, and lack of standard evaluation methods. The authors stressed the need for continued research and proposed directions to improve detection effectiveness.

Similarly, [Wu et al. 2025] reviewed recent advances in detecting AI-generated text. It highlighted several challenges, including out-of-distribution data, adversarial attacks, and variability in real-world data. The study also pointed out the lack of effective evaluation frameworks. The authors highlighted the urgent need to improve detection methods. They suggested future research should focus on developing robust detectors against attacks, creating diverse and standardized datasets, and improving generalization to unseen data. Additionally, establishing reliable evaluation protocols is crucial to enhance real-world applicability.

In another recent study examined how well frequent LLM users detect AI-generated text [Russell et al. 2025]. These "expert" annotators performed exceptionally well without any special training. Their majority vote misclassified only 1 out of 300 articles. They outperformed most commercial detectors, even under tricks like paraphrasing. Experts relied on lexical cues and deeper aspects like tone, clarity, and originality. However, the study only used American English articles and did not test factual accuracy. The authors suggest that training human annotators and combining them with tools like Pangram could further improve detection.

Previous studies have used N-gram models, classical ML, DL, and transformers to detect AI-generated text and have evaluated online tools like GPTZero and OpenAI's classifier. However, they did not combine multiple model types with diverse feature sets, nor did they systematically test lightweight models like logistic regression and DistilBERT. This study fills these gaps by integrating varied features and custom classification layers tailored to different models, and introducing an ensemble approach to improve detection accuracy and efficiency.

## 3 Methodology

### 3.1 Data Acquisition

Our study aims to compare the texts written by humans and those generated by GPT-3.5 in a research context by curating 250 pairs of research abstracts from a wide range of research themes. Due to the scarcity of abstract data collected, we created a relatively smaller labeled dataset. To ensure the human-generated abstracts were not influenced by AI-generated content, we collected abstracts published before 2010. Next, employing OpenAI's GPT-3.5 language model, we generated an equal number of artificial abstracts, resulting in a balanced dataset of 250 pairs of human-generated and AI-generated abstracts. Similarly, we created a test dataset with 200 instances, including both AI-generated and human-generated text, to test and compare the performance of the models. Later, as resources became more accessible, a labeled dataset consisting of 10,000 instances was downloaded from the Kaggle website and used.

### 3.2 Overview of the Models

Both classical ML models and transformer-based ML models were applied to perform the task of AI-generated text detection. Here, we aimed to compare their diverse methodological approaches systematically in detecting the AI-generated text. The first model employed logistic regression, incorporating BoW, POS, and TF-IDF features to improve the effectiveness of classical ML techniques. This is also to verify whether a less complex model could provide competitive results compared to more complex neural network models. The second model utilized LSTM with N-gram features to capture contextual information, aiming to explore how well an architecture specialized in handling sequences could effectively detect AI-generated text.

Transformer-based models included in this study are as follows: The third model integrated BERT with N-gram features, intending to enhance the model's contextual understanding. The fourth model featured DistilBERT, which was chosen to explore whether a distilled version of the transformer-based model could retain its effectiveness. Finally, we implemented BERT with a custom classifier to further enhance understanding of the patterns. These models were chosen to explore various aspects of AI-generated text detection, to identify the most accurate model for detecting AI-generated text based on the evaluation metrics. To explore the benefits of combining different approaches in detecting, an ensemble technique is implemented utilizing the collective strength of the 3 best-performing models. Upcoming sections provide details about the specifics of each model and its respective approaches.

### 3.3 Classical Logistic Regression ML Model

Here, logistic regression is used as a classification ML algorithm, and it integrates a combination of feature extraction techniques to enhance the model's ability to discern the patterns within the text. The detailed process is illustrated in the flowchart in Figure

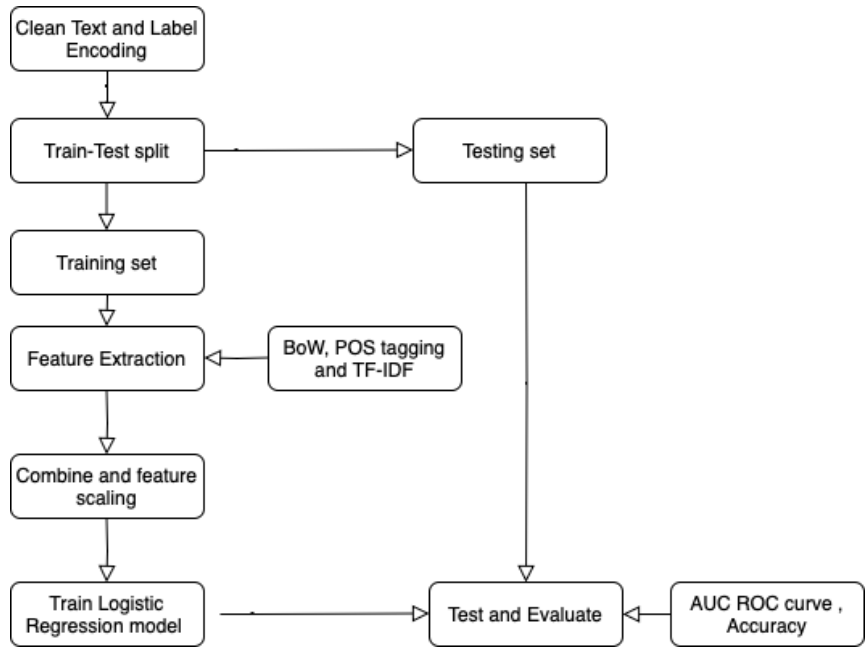

*Figure 2: Flowchart of Logistic Regression Method*

2, outlining the overall approach. Feature extraction and engineering is a crucial step in text classification using ML, and it is executed through the deployment of various techniques. The feature extraction techniques utilized for this model are as follows:

1. Bag of words (BOW) – It is a foundational approach in text analysis that represents a document as an unordered collection of words, discarding grammar and word order. Here, CountVectorizer is employed to transform the text into a numerical matrix where each row corresponds to a sentence and each column represents the frequency of a unique word.

   This captures the presence and the frequency of words, allowing the model to learn patterns based on word occurrences.

2. TF-IDF – This is a feature extraction technique that considers the importance of words by weighing them based on their frequency in a text relative to their occurrence in the entire dataset. Here Tfidf Vectorizer is utilized to convert the text into a matrix of TF-IDF features.

   It addresses the issue of common words by down-weighting them and focusing the model's attention on words that are important

3. POS tagging – To capture syntactic information and understand the grammatical structure of the text, POS tagging is incorporated. This involves assigning parts of speech(e.g., noun, verb) to each word in a sentence. The extracted POS tags are then concatenated into a string and used as additional features. Refer the Figure 3 for the distribution of POS tags in the train and test sets.

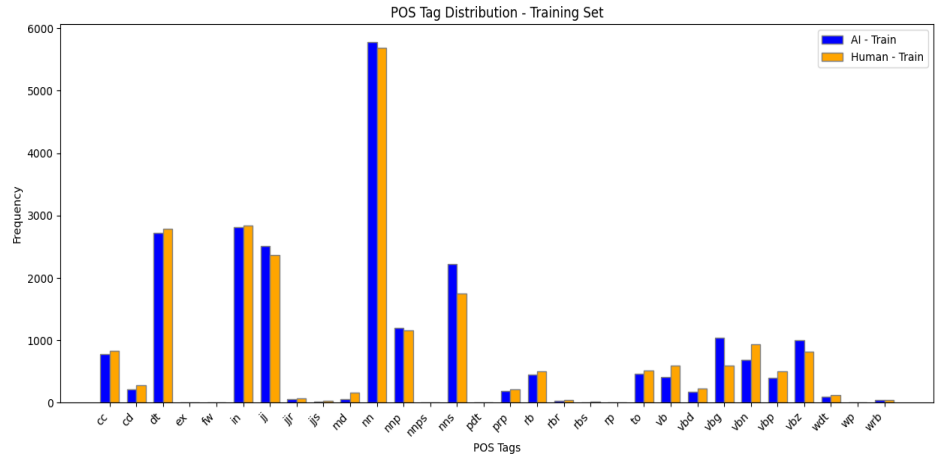

*(a)*

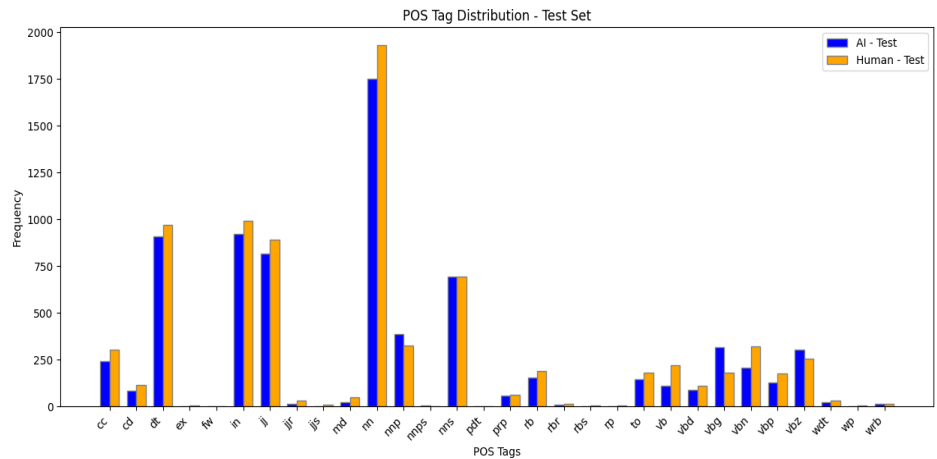

*(b)*

Figure 3: POS Distribution Comparison of (a) Train Set (b) Test Set

BOW, TF-IDF, and POS features are concatenated to form a comprehensive feature set for training the logistic regression model. Incorporating diverse feature types enables the model to learn more effective decision boundaries. The combination of features helps the model generalize across different writing styles. To ensure uniformity across feature scales and prevent any single feature from disproportionately influencing the model's learning process, we employ standard scaling. These varied pre-processing steps collectively prepare the textual data for the training and evaluation stages, ultimately enhanc-ing the model's capacity to understand the patterns and make accurate classifications. The model architecture features a linear layer with a sigmoid activation function for binary classification. The chosen binary cross-entropy loss function and stochastic gradient descent optimizer contribute to the model's efficiency. Following the training, we evaluated the model on a separate test dataset. Evaluation metrics such as AUC score and accuracy were used to assess the model's ability to distinguish between different AI-generated and human-written text.

### 3.4 Recurrent Neural Network Model: LSTM with N-gram Features

Combining LSTMs with N-gram features improves traditional neural networks by helping the model remember important information and find intricate patterns. In this approach, N-gram features are used in classifying the input text. Figure 4 shows the model's architecture and operation.

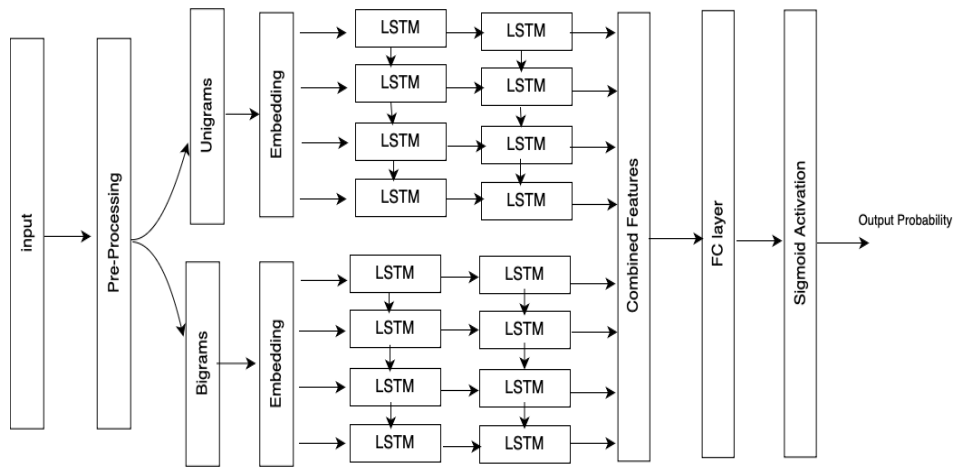

*Figure 4: Flowchart of LSTM N-gram Method*

This method preprocesses and tokenizes textual data for ML by cleaning and standardizing the input by removing non-word characters and replacing white-space, and digits with no space. It converts text into token sequences of uniform length using zero-padding to fit neural network requirements. The tokenization maps words to integer indices based on frequency and encodes labels as "AI" = 1 and "Human" = 0. The processed data and vocabulary are then prepared for training and testing. Although the initial version included a stop words removal function, it was later excluded from the final implementation. The decision to exclude stop word removal reflects the understanding that certain countries or regions may exhibit distinct linguistic patterns. For instance, some countries frequently use words like "the" while others may use them less frequently. The inclusion of such stop words may contribute valuable contextual information for text classification.

Another method is implemented to generate bigrams from the input text and returns both a list of bigrams and a counter object that counts the frequency of each bigram.

The frequency of each bigram is calculated for both training and testing datasets. The resulting bigram vocabulary is integrated with unigram features during model training. In this method, we define two embedding layers, one for unigrams and one for bigrams. These layers map the tokenized words to dense vectors of specified dimensions. Each layer is designed to capture the sequential dependency within the input data. Notably, the original model had a single LSTM layer, whereas the modified model stacks two LSTM layers consecutively. Using multiple LSTM layers in a neural network allows the model to learn hierarchical features and capture more complex relationships within the input data. In the context of text classification, the inclusion of two LSTM layers for both unigrams and bigrams provides the model with the ability to learn increasingly abstract representations of the input text. The first LSTM layer captures local dependencies within the sequence, while the second layer refines this information to capture high-level patterns. This enables the model to capture both local and high-level semantic patterns in the text.

The output representations from unigrams and bigrams are concatenated, and the dropout approach is applied to the combined representation. Integrating the unigrams and bigrams outputs enhances the model's ability to capture complex structures and different levels of linguistic patterns. This combined representation is then passed through a fully connected layer with a sigmoid activation function to produce the final output.

In this binary classification problem, the BCEWithLogitsLoss function is used because our goal is to classify inputs into two different classes. It is common in such classification tasks to use a sigmoid activation function in the output layer. BCEWithLogitsLoss combines sigmoid activation and binary cross-entropy loss, providing numerical stability during training. This loss function can be more numerically stable than manually applying a sigmoid activation followed by CrossEntropyLoss. The reason for this stability is that BCEWithLogitsLoss combines these operations in a single step, preventing intermediate instability that may occur when applying the sigmoid function separately.

Additionally, optimizers play a crucial role in training ML models, including traditional neural networks. Their primary purpose is to minimize the loss function during the training process. This method uses Adam optimizer, which is an adaptive optimization algorithm. Its adaptive learning rates are beneficial for different parameters with different learning rates. BCEWithLogitsLoss in combination with the Adam optimizer contributes to an efficient and stable training approach for binary classification models. The training process is organized into epochs, with the model updating its parameters based on batches of data within each epoch. After training the model, instead of using a fixed threshold for classification, this method computes the ROC curve and the AUC score. The optimal threshold for classification is chosen based on the ROC curve, aiming to maximize the true positive rate while minimizing the false positive rate. This threshold is then applied to the predicted probabilities to produce the final binary classification.

### 3.5 Transformer Based Models

#### 3.5.1 BERT Model with N-gram

This method combines BERT embeddings and traditional N-gram features to address the task of AI-generated text classification. BERT is a pretrained transformer-based model that captures deep contextual relationships within text, while N-gram features extract local patterns based on word sequences. By integrating the strengths of both representations, the model eventually improves its ability to understand and classify texts with enhanced accuracy.

To prepare the data for training and evaluation, the dataset is first loaded and preprocessed into a format suitable for training. BERT necessitates tokenized input; thus, we use the BERT tokenizer to segment the text into subword tokens, making it compatible with the BERT model's input requirements. Padding and truncation are applied to ensure that all the input sequences have the same length for efficient processing. Here, N-gram features capture local patterns and relationships between adjacent words, providing coherent linguistic information. We use a count vectorizer configured to extract unigrams to quadgrams from the input text. The text data is converted into a matrix of N-gram features representing the occurrence of each N-gram per input sample. The resulting N-gram feature matrices are then converted into PyTorch tensors, which are necessary to seamlessly integrate N-gram features with the BERT input.

In the context of the BERT model, an attention mask is used to indicate which elements of the input sequence should be attended to and which should be ignored. The mask is primarily used to mask out the padding tokens during attention computation, ensuring that the model focuses solely on meaningful content within the input sequence. Here, the mask is generated using the BERT tokenizer and then extended to accommodate n-gram features. This modified attention mask is applied during both the training and evaluation phases to ensure that the model appropriately considers both BERT embeddings and n-gram features.

We implement a custom model for text classification by combining a pre-trained BERT model with additional n-gram features. The output is obtained by passing the tokenized input through the BERT model, resulting in the pooled output (i.e., [CLS] token representation), which is then concatenated with the n-gram features. BERT embeddings capture the contextual information from input text, and n-gram features capture the patterns and relationships within the sequences. Concatenating them allows the model to leverage both BERT embeddings and n-gram features for better representation. The resulting tensor is passed through a linear layer to obtain the logits for classification, with ReLU activation applied to these logits. Figure 5 provides a visual representation of the workflow.

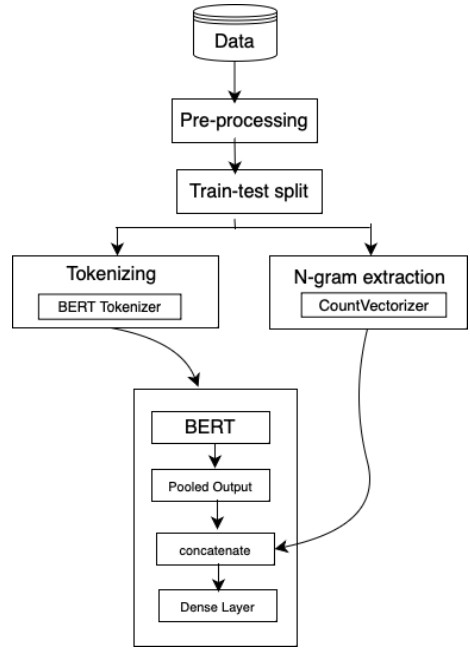

*Figure 5: BERT N-gram Approach Flowchart*

### 3.5.2 Pre-trained DistilBERT

We also utilize DistilBERT, a transformer-based model distilled from BERT, to evaluate its standalone effectiveness. This approach excludes the use of auxiliary features, allowing for direct assessment of the model's raw capabilities in distinguishing AI-generated text. Transfer learning is effective when labeled data is limited, as it allows a model to apply pre-trained knowledge to a new task. In this approach, only the classification head is updated while keeping the DistilBERT layers frozen. This enables the model to retain general language understanding while adapting to task-specific patterns. The DistilBERT tokenizer and a DistilBERT model for classification are initialized using the 'distilbert-base-uncased' pre-trained model.

Here, the input text data is first tokenized using the DistilBERT tokenizer, and then the fine-tuning process utilizes the 'DistilBertForSequenceClassification' model. Following data loading, the pre-processing steps are executed, which include removing punctuations, handling missing values, and label encoding. Since DistilBERT relies on high-quality tokenized input, effective tokenization is crucial to capture meaningful word relationships. Additionally, robust pre-processing improves the model's ability to generalize and improves its predictive performance on previously unseen data.

### 3.5.3 Pre-trained BERT Model with Custom Classifier

This method implements a transfer learning approach using BERT. The flowchart in Figure 6a illustrates the sequential steps of this approach. BERT utilizes a stack of trans-former encoders, with each encoder incorporating a self-attention mechanism that enables it to weigh the importance of different words in a sequence based on context. In this method, the parameters of the pre-trained model are frozen, and it is employed as a fixed feature extractor. BERT performs feature extraction, followed by a custom classifier built with PyTorch neural network modules. Its multi-layer architecture refines token representations through self-attention and feedforward layers. As the input sequence traverses these layers, tokens are refined based on their contextual information.

The custom classifier layers, which include a dropout layer, ReLu activation, two dense layers (fully connected), and SoftMax activation as shown in Figure 6b, are used. The dropout approach is employed to reduce overfitting during the training phase by setting the 10% input units to zero at each update during training. Subsequently, ReLU activation is applied, enabling the model to learn complex patterns and relationships within the data. The first dense layer transforms the BERT hidden size (768) into an intermediate feature size (512), while the second dense layer maps these intermediate features to the output classes. This output layer produces the final output logits for each class, preparing the model for classification.

During the forward pass, the BERT-encoded input goes through the custom classifier, which applies a softmax activation function to generate class probabilities. This process involves extracting the representation of [CLS] tokens from the BERT output, passing it through the custom classifier layers, and ultimately transforming it into class probabilities using softmax activation.

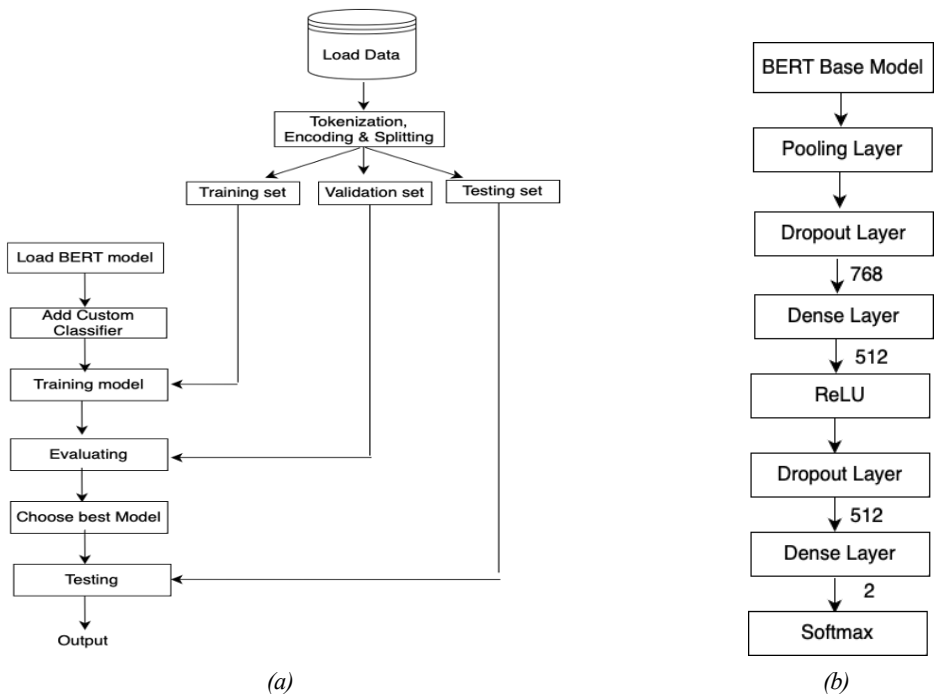

*Figure 6: BERT with Custom Classifier Approach (a)Flowchart of the Approach, (b) Custom Classifier Layers*

# 4 Results and Discussion

The evaluation of each approach's effectiveness is assessed using crucial metrics such as ROC curves, classification reports, and loss curves. These metrics aid in understanding the strengths and weaknesses of each model. It is essential to highlight that the models initially underwent training with the specifically created dataset, followed by an evaluation of their performance using an external test dataset. Furthermore, two out of the five models were trained with the downloaded dataset. Additionally, a max voting ensemble technique was implemented using the three best-performing models. The following subsections explore each model's performance in more detail.

## 4.1 Performance Comparison of the Models

The logistic regression model achieved an accuracy of 85.57% as mentioned in Table 1, correctly classifying instances approximately 85.57% of the time. The recall metric illustrates that the model captures approximately 86% of true AI instances. Furthermore, the F1-score highlights the model's effective balance between making accurate positive predictions and correctly capturing instances. Overall, this model showcases a strong capability in distinguishing AI-generated text from human-written content.

In LSTM with the N-gram model, several crucial observations have significantly influenced the model's performance. Various adjustments made to the model have resulted in an enhancement in accuracy, elevating the performance from an initial 47% to 57%. However, initially, the removal of stop words led to a decrease in accuracy, suggesting that these common words may contain valuable information for the prediction task. The switch from the entropy Loss function to BCEWithLogits further enhanced the model's accuracy. Additionally, dynamically selecting the classification threshold using the ROC curve instead of a fixed 0.5 threshold contributed to better performance. These observations highlight the importance of the parameter choices to achieve optimal performance. This model has an accuracy of 57%. For LSTM with N-gram, precision for the "AI" class is 0.60, indicating reasonable accuracy when identifying AI-generated text, but recall is relatively low at 0.44, meaning a substantial portion of actual AI instances are missed.

According to the classification report in Table 1, the BERT with N-gram model achieved an accuracy of 91%, demonstrating the strongest overall predictive performance. This high accuracy indicates effective discrimination between classes, with balanced accuracy suggesting minimal bias across categories. Both precision and recall are high for each class, indicating that the model is effective in correctly identifying instances of both classes.

The DistilBERT model also showed strong performance with an impressive accuracy of 90% on the created dataset. It further shows high precision, recall, and F1 score for both classes, indicating balanced classification. The classification report illustrates the effectiveness of the model with high macro- and weighted average in precision, recall, and F1 score, suggesting an effective handling of class imbalances. Consistent scores across classes underscore the model's capability to manage the dataset's complexities.

The BERT with Custom Classifier model exhibits a reasonable level of performance, achieving an accuracy of 75%. Its precision, measuring the accuracy of positive predictions, stands at 70.21% indicating that nearly 70% of the instances predicted as AI were indeed AI. This model maintains a good balance between precision and recall, making it effective for this classification task.

In summary, (Table 1) DistilBERT model stands out as a robust choice. However, the Logistic regression model and BERT with N-gram also demonstrate strong overall performance and may be suitable in some cases.

| Models | Accuracy | Precision | F1-score | Recall |
|---|---|---|---|---|
| Logistic Regression Model | 85.57% | 0.86 | 0.86 | 0.86 |
| LSTM N-gram | 57% | 0.60 | 0.51 | 0.44 |
| BERT N-gram | 90.66% | 0.96 | 0.91 | 0.87 |
| BERT Custom Classifier | 75% | 0.82 | 0.71 | 0.62 |
| DistilBERT | 90% | 0.98 | 0.90 | 0.83 |

*Table 1: Classification Report of the Models*

The ROC curve is used to evaluate model performance by illustrating the trade-off between sensitivity and specificity. This section presents the ROC-based performance comparison of the models.

The AUC of the Logistic regression model appears to be 0.86 as shown in Figure 7a, indicating a decent performance in distinguishing text. The ROC curve rises steadily, though less steeply at lower true positive rates, suggesting room for improvement in early-stage classification.

According to Figure 7b the area under the curve(AUC) score of the DistilBERT model appears to be 0.90, indicating very good performance in distinguishing AI-generated text. The smooth, steeply rising curve at low false positive rates suggests the model effectively captures true positives while minimizing false detections. Figure 7c shows that the LSTM N-gram model has a slightly lower AUC score. This model also has a less smooth curve indicating room for improvement. BERT with N-gram as shown in Figure 7d steeply rises at low positive rates, indicating high sensitivity. It's AUC of 0.912 suggests that the model has a strong ability to distinguish between positive and negative classes. According to Figure 7e, the BERT with Custom Classifier model achieves an AUC score of 0.745, indicating a reasonable ability to distinguish between AI-generated and human-written text.

In the comparative analysis, DistilBERT and BERT with N-gram show the highest AUC scores, indicating strong performance in distinguishing AI-generated text. They are followed by LSTM N-gram and BERT with Custom Classifier, with slightly lower AUC values, while Logistic Regression records the lowest. Although most ROC curves appear smooth, suggesting effective learning, those for the BERT with Custom Classifier and Logistic Regression models show less smoothness, indicating potential areas for improvement.

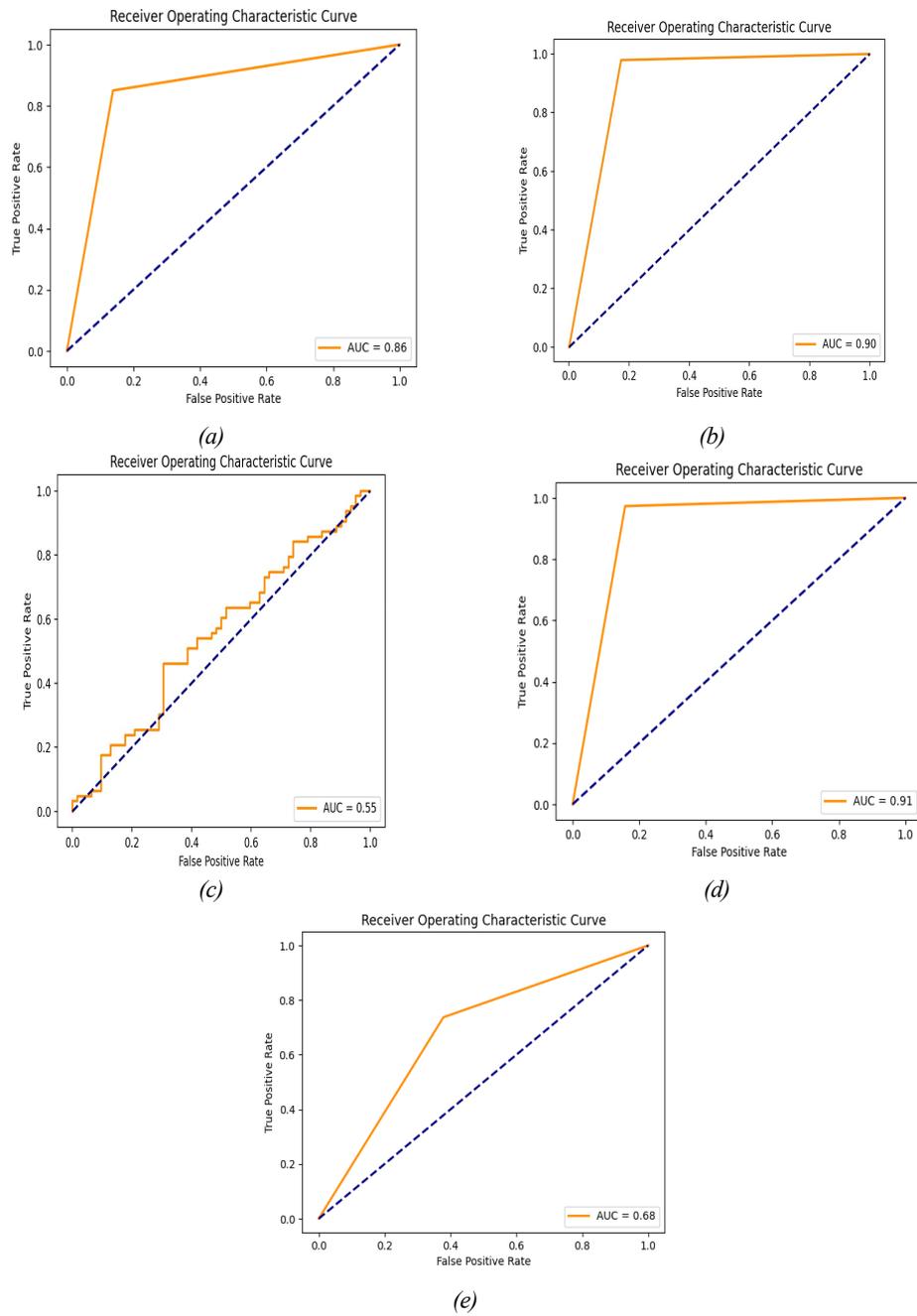

*Figure 7: ROC Curves of Different Models (a) Logistic Regression, (b) DistilBERT, (c) LSTM with N-gram, (d) BERT with N-gram, (e) BERT with Custom Classifier*

## 4.2  Behavior of Loss Curves

In these models, understanding the training and validation performances is crucial for assessing the model's overall performance systematically. The visual representation of these processes is the loss curves. Comparing loss curves of training and validation gives insights into both accuracy as well as generalizability. The loss graphs in Figure 8 illustrate the evolution of the model's loss over epochs during the validation and training.

According to the Figure 8a, the loss curves of the Logistic Regression model for both training and validation steadily decrease over the first 6 epochs, suggesting effective learning. Beyond this point, the training loss continues to decline slightly while the validation loss stabilizes. This suggests limited generalization capability. In the BERT N-gram model, the validation loss diverges slightly from the training loss as illustrated in Figure 8b, though the gap remains minimal. Notably, the validation loss consistently remains lower than the training loss, which is generally a positive sign.

As shown in Figure 8c, the training loss of the DistilBERT model continues to decline while the validation loss stabilizes. The validation loss doesn't diverge substantially from the training loss, suggesting that the model's performance on unseen data reaches a point of limited improvement. Figure 8d shows the loss graph of BERT with a custom classifier. Both the training and validation losses drop sharply within the first few epochs, indicating fast learning. However, the validation loss suggests that there's room for potential improvement in the model's generalization performance. As shown in Figure 8e, the LSTM N-gram model's training loss begins high and decreases rapidly, indicating effective learning. The slower decline in validation loss suggests good generalization to unseen data.

Across all loss graphs, BERT with N-gram shows rapid initial learning, while DistilBERT demonstrates steady convergence and slightly lower validation loss, indicating better generalization. Logistic Regression exhibits higher validation loss compared to BERT-based models indicating poor learning performance. Although BERT with a Custom Classifier also shows fast learning, the overall comparison suggests DistilBERT as the most balanced and reliable model.

Delving into the comparative analysis, DistilBERT and BERT with N-gram appear to have the highest AUC scores, indicating excellent performance in distinguishing AI-generated text. They are followed by LSTM N-gram and BERT with a custom classifier, followed by a slightly lower AUC score, while Logistic regression records the lowest AUC. Most curves appear smooth, reflecting effective learning. However, BERT with a custom classifier and Logistic regression show less smooth curves, suggesting room for improvement.

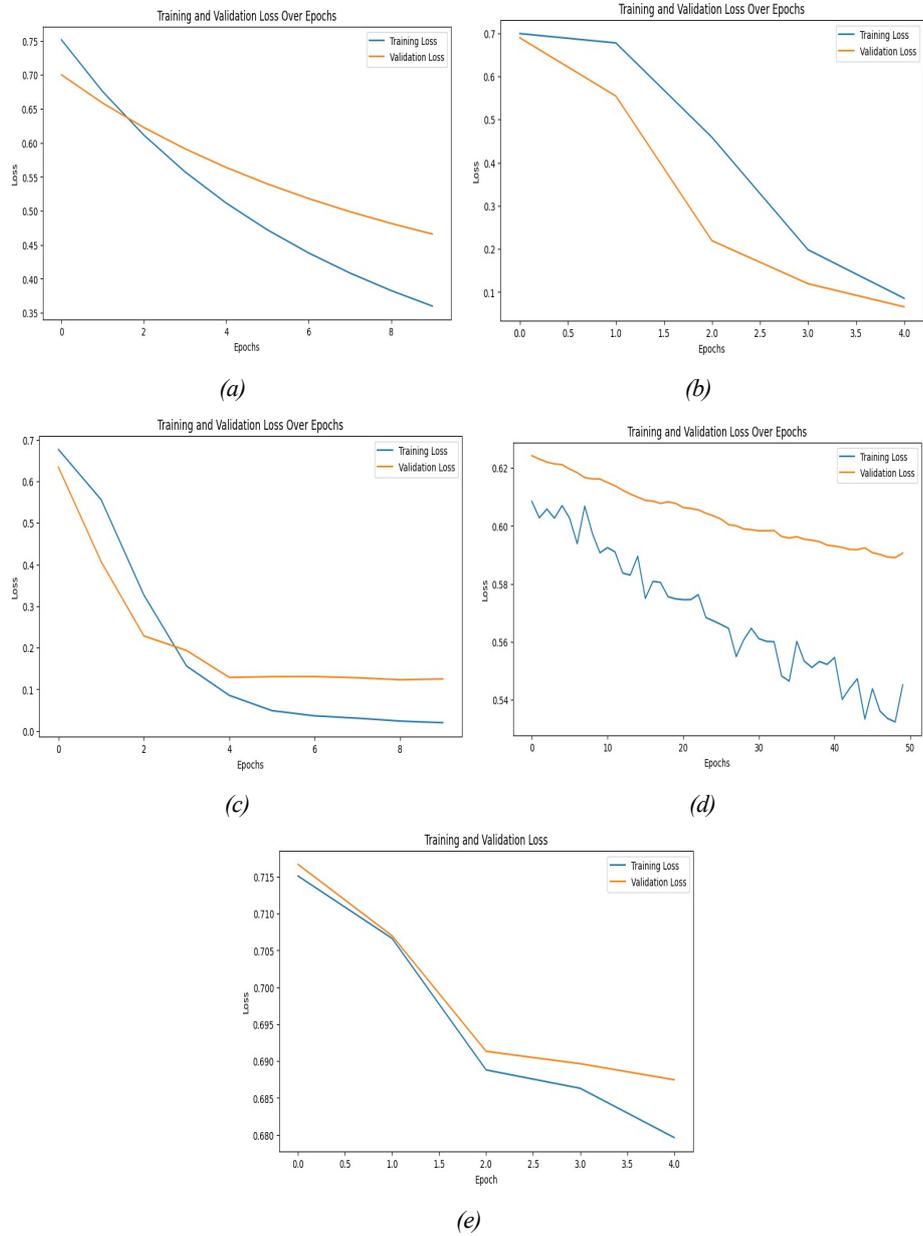

*(e)*

*Figure 8: Loss Graphs of Different Models (a) Logistic Regression, (b) BERT N-gram, (c) DistilBERT, (d) BERT with Custom Classifier, (e) LSTM N-gram*

**4.3  Performance Validation on External Test Data**

This section presents the evaluation of model performance using a dedicated test set, which is completely outside the training data to ensure an unbiased assessment. This will give insights into the model's real-world applicability after comprehensive training and validation.

As shown in Table 2, DistilBERT emerges as the top-performing model with an accuracy of 90.54%. This high accuracy is indicative of the model's ability to make correct predictions across both AI and Human classes. Furthermore, the model demonstrates high precision, confirming that its predictions are reliably correct. Following closely is the Logistic Regression Model showing a solid accuracy of 79.54%. While its precision values are relatively strong, the recall for AI is notably low. In comparison, LSTM N-gram and BERT N-gram models demonstrate lower overall accuracy, precision, and F1-score. These models face challenges in correctly identifying instances from the test set, possibly due to limitations in capturing complex patterns and dependencies within data.

In addition to the table, Figure 9 provides a visual representation of the classification metrics for all five models. This graphical overview offers a complete insight into the comparative strengths and weaknesses of each model, enhancing the understanding of their performance on the dedicated test set.

| Models | Accuracy | Precision | F1-score | Recall |
| --- | --- | --- | --- | --- |
| Logistic Regression Model | 79.54% | 0.84 | 0.79 | 0.74 |
| LSTM N-gram | 45.72% | 0.47 | 0.56 | 0.68 |
| BERT Custom Classifier | 76.38% | 0.77 | 0.76 | 0.75 |
| BERT N-gram | 53.26% | 0.57 | 0.42 | 0.34 |
| DistilBERT | 90.54% | 0.98 | 0.91 | 0.84 |

*Table 2: Classification Report of testing*

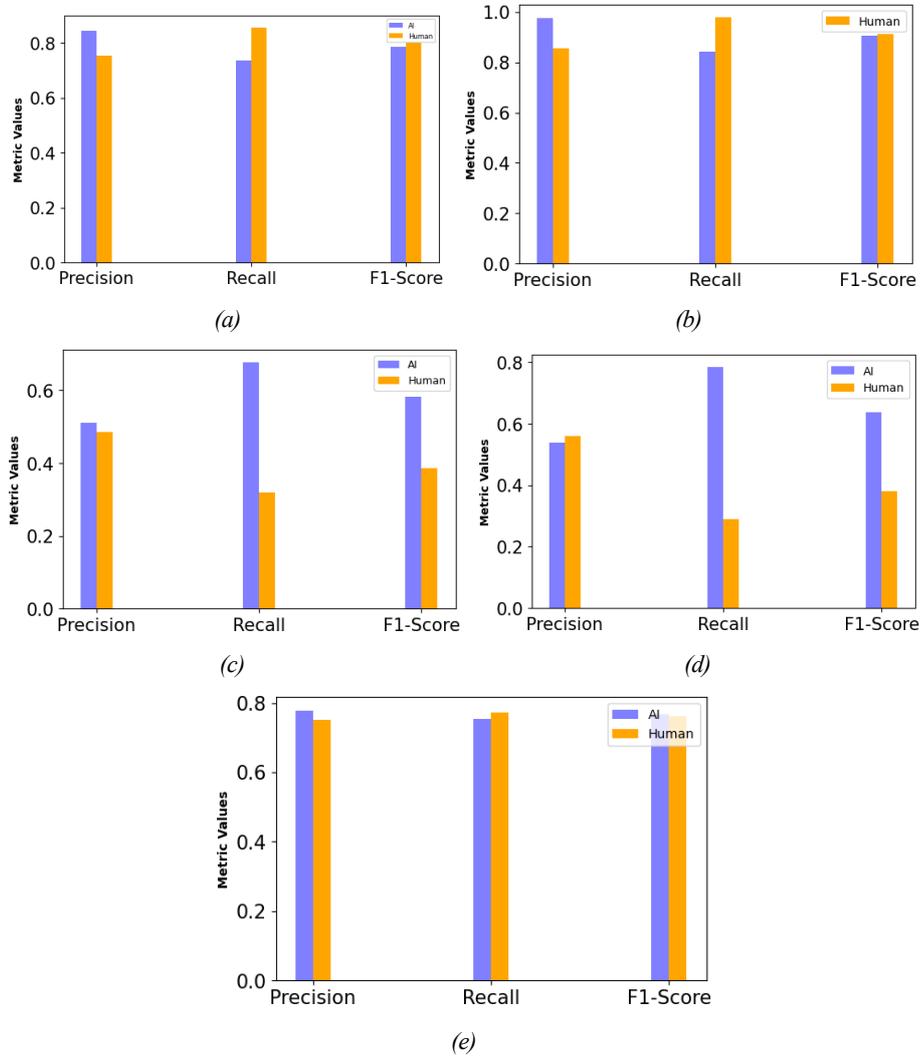

*Figure 9: Classification Performance Metrics of (a) Logistic Regression, (b) DistilBERT, (c) LSTM with N-gram, (d) BERT with N-gram, (e) BERT with Custom Classifier on External data*

### 4.4 Ensemble Classifier

Among all five trained models, we selected the three best-performing models and opted for the max voting ensemble technique to boost performance, as shown in Figure 10. This technique utilizes the strength of individual models while reducing the impact of potential weaknesses. The predictions from the best three models, DistilBERT, BERT Custom Classifier, and Logistic Regression, are passed through a max voting classifier to generate the final prediction, which achieved an accuracy of 84.422%. Note that, for this case, the accuracy of the DistilBERT model alone surpassed the performance of the ensemble technique.

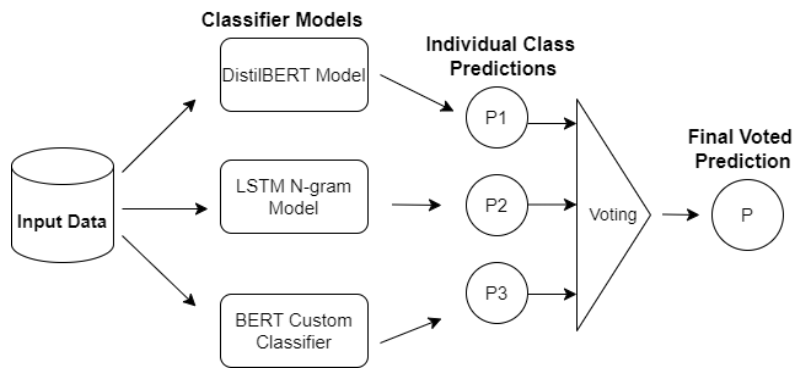

*Figure 10: Max-Voting Ensemble Classifier*

### 4.5 Evaluation with Kaggle Data

We trained and evaluated the BERT with a custom classifier and DistilBERT models using the downloaded Kaggle dataset. Importantly, as DistilBERT emerged as the best model, its performance was further assessed on this downloaded dataset to validate its robustness. The following analysis provides a detailed account of the evaluation metrics.

According to Table 3, the BERT Custom Classifier shows moderate performance. Its overall accuracy is around 58%. The recall for AI-generated content is 73%, meaning it correctly identifies most AI instances. However, there are noticeable differences in precision, recall, and F1-score between AI and Human classes. Overall, the model performs moderately in distinguishing the two types of text.

DistilBERT model achieves an accuracy of 88.65%, indicating strong overall classification performance. Precision for both AI and Human classes is high, around 88-80%, showing reliable predictions. The F1 scores are also balanced, reflecting a strong compromise between precision and recall. Overall, the model exhibits an excellent performance.

| Models | Accuracy | Precision | F1-score | Recall |
|---|---|---|---|---|
| BERT Custom Classifier | 58.02% | 0.56 | 0.63 | 0.73 |
| DistilBERT | 88.65% | 0.88 | 0.89 | 0.90 |

*Table 3: Classification Report of the Models*

## 5 Conclusion and Future Work

The present study aimed to evaluate the efficacy of various approaches in detecting the text generated by the GPT3.5 chat model. It evaluates the performance of Logistic Regression with Bag of Words, Part of Speech, and TF-IDF features; Bert with N-gram features; DistilBERT; Bert with Custom Classifier; and LSTM N-gram models in detecting AI-generated text. The result of this study indicates considerable variability in the models' ability to correctly identify and categorize AI-generated or human-written text. Notably, the varying performance underscores the complexities involved in distinguishing between AI and human-generated text.

However, DistilBERT demonstrated the highest accuracy in detecting AI-generated text, while Logistic Regression and BERT with Custom Classifier also showed commendable performance with balanced accuracy. In contrast, LSTM N-gram and BERT N-gram models exhibited challenges in capturing the intricate patterns required, resulting in comparatively lower effectiveness. An ensemble technique was introduced to further enhance the overall detection performance of LSTM N-gram and BERT N-gram. This ensemble classifier architecture consisted of two main steps. First, model selection: shortlisting three out of the five initially tested models based on their superior performance metrics. Second, an ensemble methodology employed max voting to combine the predictions of these three models to enhance overall accuracy and robustness. Despite the theoretical advantages, the results indicated that the ensemble classifier eventually did not outperform the best individual model, DistilBERT. Specifically, the accuracy of the ensemble model was lower than that achieved by DistilBERT alone, suggesting that the added complexity of combining models using max voting did not translate into better predictive performance.

These findings contribute not only to the selection of an optimal model for the task but also offer valuable insights for future research and model refinement towards AI-generated text detection. As this field continues to evolve, this study provides a meaningful contribution to understanding fundamental behaviours of different models for AI text detection and establishing the way for more sophisticated and effective solutions.

Future work could involve the integration of advanced transformer models with a hierarchical architecture. Because these models are capable of learning hierarchical representations, enabling them to capture patterns and contexts in AI-generated and human-written text, improving the detection capacity. It is of note that the current study utilized a limited-sized dataset for training and testing. Expansion of the dataset may lead to further enhancement of the model's learning capabilities and overall effectiveness.

# References

[Kreps et al. 2022] Kreps, S., McCain, R.M. and Brundage, M.,: "All the news that's fit to fabricate: AI-generated text as a tool of media misinformation."; Journal of experimental political science, 2022. [Online]. Available: https://doi.org/10.1017/XPS.2020.37.

[Woo et al. 2023] Woo, D.J., Susanto, H., Yeung, C.H., Guo, K. and Fung, A.K.Y.: "Exploring AI-Generated text in student writing: How does AI help?"; Language Learning & Technology, 2023. [Online]. Available: https://www.lltjournal.org/item/1127/.

[Partadiredja et al. 2020] Partadiredja, Arkan, R., and Entrena-Serrano, Carlos, Ljubenkov, Davor: "AI or human: the socio-ethical implications of AI-generated media content"; 13th CMI Conference on Cybersecurity and Privacy, 2020. [Online]. Available: https://ieeexplore.ieee.org/abstract/document/9322673.

[Amirjalili et al. 2024] Amirjalili F, Neysani M, Nikbakht: "A comparative analysis of AI-generated text and human academic writing in English literature"; Frontiers in Education, 2024. [Online]. Available: https://www.frontiersin.org/journals/education/articles/10.3389/feduc.2024.1347421/full.

[Wei et al. 2024] Wei, Y. and Tyson, G.,: "Understanding the impact of ai-generated content on social media: The pixiv case"; Proceedings of the 32nd ACM International Conference on Multimedia, 2024. [Online]. Available: https://doi.org/10.1145/3664647.3680631.

[Ouyang et al. 2022] Ouyang L., Wu, J., Jiang, X., Almeida, D., Wainwright, C., Mishkin, P., Zhang, C., Agarwal, S., Slama, K., Ray, A. and Schulman, J.: "Training language models to follow instructions with human feedback"; Advances in neural information processing systems (NeuralIPS Proceedings), 2022.

[Wang et al. 2023] Wang, Y., Pan, Y., Yan, M., Su, Z. and Luan, T.H.: "AI–generated contents, challenges, and solutions"; IEEE Open Journal of the Computer Society, 2023. [Online]. Available: https://ieeexplore.ieee.org/abstract/document/10221755.

[Faouzi et al. 2023] Faouzi, J. and Colliot, O.: "Classic machine learning methods";Machine learning for brain disorders, 2023. [Online]. Available: https://link.springer.com/protocol/10.1007/978-1-0716-3195-9_2.

[Nusinovici et al. 2020] Nusinovici, S., Tham, Y.C., Yan, M.Y.C., Ting, D.S.W., Li, J., Sabanayagam, C., Wong, T.Y. and Cheng, C.Y.: "Logistic regression was as good as machine learning for predicting major chronic diseases"; Journal of clinical epidemiology, 2020. [Online]. Available: https://doi.org/10.1016/j.jclinepi.2020.03.002.

[Amatriain et al. 2023] Amatriain, X., Sankar, A., Bing, J., Bodigutla, P.K., Hazen, T.J. and Kazi, M.: " Transformer models: an introduction and catalog"; arXiv preprint arXiv:2302.07730, 2023.

[Gehrmann et al. 2019] Gehrmann, S., Strobelt, H. and Rush, A.M.: " Gltr: Statistical detection and visualization of generated text"; arXiv preprint arXiv:1906.04043, 2019.

[Wu et al. 2023] Wu, K., Pang, L., Shen, H., Cheng, X. and Chua, T.S.: " LLMDet: A third party large language models generated text detection tool"; arXiv preprint arXiv:2305.15004., 2023.

[Alamleh et al. 2023] Alamleh, H., AlQahtani, A.A.S. and ElSaid, A.: "Distinguishing human-written and ChatGPT-generated text using machine learning"; Systems and Information Engineering Design Symposium (SIEDS), 2023. https://doi.org/10.1109/SIEDS58326.2023.10137767.

[Lai et al. 2024] Lai, Z., Zhang, X. and Chen, S.: "Adaptive ensembles of fine-tuned transformers for llm-generated text detection"; International Joint Conference on Neural Networks (IJCNN), 2024. https://doi.org/10.1109/IJCNN60899.2024.10651296.

[Antoun et al. 2023] Antoun, W., Mouilleron, V., Sagot, B. and Seddah, D.: "Towards a robust detection of language model generated text: is ChatGPT that easy to detect?"; arXiv preprint arXiv:2306.05871, 2024.

[Mikros et al. 2023] Mikros, G.K., Koursaris, A., Bilianos, D. and Markopoulos, G.: "AI-Writing Detection Using an Ensemble of Transformers and Stylometric Features"; IberLEF@ SEPLN, 2023.

[Hazim et al. 2024] Hazim, L.R. and Ata, O.: "Textual Authenticity in the AI Era: Evaluating BERT and RoBERTa with Logistic Regression and Neural Networks for Text Classification"; International Symposium on Electronics and Telecommunications (ISETC), IEEE, 2024. https://doi.org/10.1109/ISETC63109.2024.10797291

[Alhayan et al. 2024] Alhayan, F. and Himdi, H.: " Ensemble learning approach for distinguishing human and computer-generated Arabic reviews"; PeerJ Computer Science, 2024. https://peerj.com/articles/cs-2345/.

[Abburi et al. 2025] Abburi, H., Bhattacharya, S., Bowen, E. and Pudota, N.: " AI-generated Text Detection: A Multifaceted Approach to Binary and Multiclass Classification"; arXiv preprint arXiv:2505.11550, 2025.

[Lo et al. 2023] Lo, C.K.: "What is the impact of ChatGPT on education? A rapid review of the literature"; Education sciences, 2023. [Online]. Available: https://www.mdpi.com/2227-7102/13/4/410.

[García et al. 2021] García, M., Maldonado, S. and Vairetti, C.: "Efficient n-gram construction for text categorization using feature selection techniques."; Intelligent Data Analysis, 2021. [Online]. Available: https://doi.org/10.3233/IDA-205154.
García, M., Maldonado, S. and Vairetti, C., 2021. Efficient n-gram construction for text categorization using feature selection techniques. Intelligent Data Analysis, 25(3), pp.509-525.

[Suzuki et al. 2019] Suzuki, M., Itoh, N., Nagano, T., Kurata, G., Thomas, S.: "Improvements to N-gram Language Model Using Text Generated from Neural Language Model"; ICASSP 2019 - 2019 IEEE International Conference on Acoustics, Speech and Signal Processing (ICASSP), pp. 7245-7249, 2019. doi:10.1109/ICASSP.2019.8683481

[Sazli 2006] Sazli, M.: "A brief review of feed-forward neural networks"; Communications Faculty Of Science University of Ankara, vol. 50, pp. 11-17, Jan. 2006. doi:10.1501/commua1-2$_0$000000026

[Vaswani et al. 2023] Vaswani, A., Shazeer, N., Parmar, N., Uszkoreit, J., Jones, L., Gomez, A. N., Kaiser, L., Polosukhin, I.: "Attention Is All You Need"; arXiv e-print 1706.03762, cs.CL, 2023.

[Majumder et al. 2002] Majumder, P., Mitra, M., Chaudhuri, B. B.: "N-gram: a language-independent approach to IR and NLP"; International Conference on Universal Knowledge and Language, vol. 2, 2002.

[Devlin et al. 2019] Devlin, J., Chang, M. W., Lee, K., Toutanova, K.: "BERT: Pre-training of Deep Bidirectional Transformers for Language Understanding"; arXiv e-print 1810.04805, cs.CL, 2019.

[Sanh et al. 2020] Sanh, V., Debut, L., Chaumond, J., Wolf, T.: "DistilBERT, a distilled version of BERT: smaller, faster, cheaper and lighter"; arXiv e-print 1910.01108, cs.CL, 2020.

[Sarzaeim et al. 2023] Sarzaeim, P., Doshi, A., Mahmoud, Q.: "A Framework for Detecting AI-Generated Text in Research Publications"; Proceedings of the International Conference on Advanced Technologies, vol. 11, pp. 121-127, Sep. 2023. [Online]. Available: https://proceedings.icatsconf.org/conf/index.php/ICAT/article/view/36. doi:10.58190/icat.2023.28

[Elkhatat et al. 2023] Elkhatat, A. M., Elsaid, K., Almeer, S.: "Evaluating the efficacy of AI content detection tools in differentiating between human and AI-generated text"; International Journal for Educational Integrity, vol. 19, no. 1, p. 17, 2023, Springer.

[Verma et al. 2023] Verma, V., Fleisig, E., Tomlin, N. and Klein, D.: Detecting text ghostwritten by large language models."; arXiv preprint arXiv:2305.15047., vol. 19, no. 1, p. 17, 2023, Springer.

[Wang et al. 2023] Wang, P., Li, L., Ren, K., Jiang, B., Zhang, D., Qiu, X.: "SeqXGPT: Sentence-Level AI-Generated Text Detection"; arXiv e-print 2310.08903, cs.CL, 2023.

[Yan et al. 2023] Yan, D., Fauss, M., Hao, J., Cui, W.: "Detection of AI-generated essays in writing assessment"; Psychological Testing and Assessment Modeling, vol. 65, no. 2, pp. 125-144, 2023.

[Lund 2023] Lund, B.: "A brief review of ChatGPT: Its value and the underlying GPT technology"; Preprint. University of North Texas. Project: ChatGPT and Its Impact on Academia. Doi, vol. 10, 2023.

[Staudemeyer et al. 2019] Staudemeyer, R. C., Morris, E. R.: "Understanding LSTM – a tutorial into Long Short-Term Memory Recurrent Neural Networks"; arXiv e-print 1909.09586, cs.NE, 2019.

[Mitchell et al. 2023] Mitchell, E., Lee, Y., Khazatsky, A., Manning, C. D., Finn, C.: "DetectGPT: Zero-Shot Machine-Generated Text Detection using Probability Curvature"; arXiv e-print 2301.11305, cs.CL, 2023.

[Katib et al. 2023] Katib, I., Assiri, F. Y., Abdushkour, H. A., Hamed, D., Ragab, M.: "Differentiating Chat Generative Pretrained Transformer from Humans: Detecting ChatGPT-Generated Text and Human Text Using Machine Learning"; Mathematics, vol. 11, no. 15, article no. 3400, 2023. [Online]. Available: https://www.mdpi.com/2227-7390/11/15/3400. doi:10.3390/math11153400

[Fraser et al. 2025] Fraser, K. C., Dawkins, H., Kiritchenko, S.: "Detecting AI-Generated Text: Factors Influencing Detectability with Current Methods"; Journal of Artificial Intelligence Research, vol. 82, pp. 2233–2278, 2025. [Online]. Available: http://dx.doi.org/10.1613/jair.1.16665. doi:10.1613/jair.1.16665

[Wu and Segura-Bedmar 2025] Wu, L. Y., and Segura-Bedmar, I.: "AI-generated Text Detection with a GLTR-based Approach," 2025. [Online]. Available: https://arxiv.org/abs/2502.12064.

[Wu et al. 2025] Wu, J., Yang, S., Zhan, R., Yuan, Y., Chao, L. S., Wong, D. F.: "A Survey on LLM-Generated Text Detection: Necessity, Methods, and Future Directions"; Computational Linguistics, vol. 51, no. 1, pp. 275–338, 2025. [Online]. Available: https://aclanthology.org/2025.cl-1.8/. doi:10.1162/coli_{ao}0549

[Russell et al. 2025] Russell, J., Karpinska, M., Iyyer, M.: "People who frequently use ChatGPT for writing tasks are accurate and robust detectors of AI-generated text"; arXiv preprint arXiv:2501.15654, 2025. [Online]. Available: https://arxiv.org/abs/2501.15654.